\documentclass[11pt]{article}
\usepackage[ruled,vlined]{algorithm2e}
\usepackage{amsfonts}
\usepackage{amsmath}
\usepackage{amssymb}
\usepackage{authblk}
\usepackage{booktabs}
\usepackage[margin=1in]{geometry}
\usepackage{graphicx}
\usepackage{multirow}
\usepackage{xcolor}
\usepackage[colorlinks=true,citecolor=blue]{hyperref}

\title{One-shot learning for solution operators of partial differential equations}

\author[1]{Anran Jiao}
\author[2]{Haiyang He}
\author[2]{Rishikesh Ranade}
\author[2]{Jay Pathak}
\author[1,3,*]{Lu Lu}

\affil[1]{Department of Statistics and Data Science, Yale University, New Haven, CT 06511, USA}
\affil[2]{Ansys Inc., San Jose, CA 95134, USA}
\affil[3]{Wu Tsai Institute, Yale University, New Haven, CT 06510, USA}
\affil[*]{Corresponding author. Email: lu.lu@yale.edu}
\date{}

\begin{document}

\maketitle

\begin{abstract}

Learning and solving governing equations of a physical system, represented by partial differential equations (PDEs), from data is a central challenge in a variety of areas of science and engineering.
Traditional numerical methods for solving PDEs can be computationally expensive for complex systems and require the complete PDEs of the physical system. On the other hand, current data-driven machine learning methods require a large amount of data to learn a surrogate model of the PDE solution operator, which could be impractical. Here, we propose the first solution operator learning method that only requires one PDE solution, i.e., one-shot learning. By leveraging the principle of locality of PDEs, we consider small local domains instead of the entire computational domain and define a local solution operator. The local solution operator is then trained using a neural network, and utilized to predict the solution of a new input function via mesh-based fixed-point iteration (FPI), meshfree local-solution-operator informed neural network (LOINN) or local-solution-operator informed neural network with correction (cLOINN). We test our method on diverse PDEs, including linear or nonlinear PDEs, PDEs defined on complex geometries, and PDE systems, demonstrating the effectiveness and generalization capabilities of our method across these varied scenarios.

\end{abstract}

\section{Introduction}
\label{sec:intro}

Learning and solving governing equations of a physical system from data is a central challenge in a variety of areas of science and engineering. These governing equations are usually represented by partial differential equations (PDEs).
In real-world applications, however, most PDEs lack analytical solutions. Consequently, various approaches on solving PDEs have been developed over the years.
Traditional numerical methods including finite element method, finite difference method, and spectral methods have been well established. However, the computational costs of these numerical methods are prohibitively expensive for complex systems. Moreover, these methods typically require spatial discretization on some mesh of nodes, and evaluation of the solution at any other point requires interpolation or some other reconstruction method. More importantly, numerical methods require a complete understanding of the underlying PDEs.

Recently, physics-informed machine learning (PIML) has obtained great attention and provides alternative approaches for solving PDEs or approximating solution operators of PDEs~\cite{karniadakis2021physics}. By integrating physics into the loss function of a neural network using automatic differentiation, physics-informed neural networks (PINNs) have been successfully applied to solve forward problems of various types of PDEs across multiple fields~\cite{raissi2019physics,lu2021deepxde, pang2019fpinns, zhang2019quantifying, rao2020physics, wu2018physics, qian2020lift, costabal2020biomed, yu2021gradient,wang2023learning}. PINN is able to generate differentiable PDE solutions at any point without a mesh. However, solving a new PDE requires the training of a new neural network, which is still computationally expensive. To address this issue, deep neural operators have been developed very recently to learn PDE solution operators by using neural networks, such as DeepONet~\cite{lu2021learning, wang2021learning, lu2022comprehensive,jin2022mionet, zhu2023fourier, jiang2023fourier, zhu2023reliable} and Fourier neural operator~\cite{li2020fourier,lu2022comprehensive}. Deep neural operators enable fast prediction of PDE solutions under varying conditions, such as different boundary and initial conditions.

Training deep neural operators requires large amounts of data, which is either experimental data or data from computational simulations of physical systems~\cite{baker2019workshop}. However, in practice, it can be expensive or infeasible to obtain or store large amounts of such data. For instance, in geophysics applications, it is quite expensive to measure seismic activity, resistivity, ground penetrating radar, and magnetic fields~\cite{kafadar2016computer}. Similarly, in climate modeling, observed climate records are insufficient and running a long-term high-resolution climate simulation is not feasible with current computational power~\cite{balaji2021climbing}. The field of fluid dynamics faces analogous challenges, as the complexity of simulating fluid flow and heat transfer in various engineering applications requires high-performance computing resources, which is economically and computationally expensive. Hence, it would be highly beneficial if we develop methods aiming to minimize the amount of training data and the associated computational costs for learning PDE solution operators. This motivates our focus on an intriguing scenario: one-shot learning, which involves network training based on only a single data point.

One-shot learning methods have been proposed to learn from a single example usually by utilizing previously acquired knowledge~\cite{fei2006one,lake2011one,vinyals2016matching}. One-shot learning is mainly used in computer vision field and its usage in PIML is very limited~\cite{ivanov2020physics, darcy2023one}. To our knowledge, no one-shot learning method has been developed for learning PDE solution operators. In this work, we propose the first one-shot learning approach to learn PDE solution operators from only one data point. In our method, we abstract some ``general knowledge'' from only one PDE configuration and its solution, and make it possible to predict PDE solutions with other parameters or conditions.

From another perspective, our method can also be viewed as an approach for discovering PDEs. When discovering PDEs, in one scenario, where we know all the terms of the PDE and only need to infer unknown coefficients from data, many neural network-based methods have been proposed. For example, we can enforce physics-based constraints to train neural networks that can solve inverse problems of PDEs~\cite{lu2021deepxde, chen2020physics, zhang2019quantifying, yazdani2020systems, lu2021physics, daneker2023systems, tartakovsky2020fluid, wu2022effective}. In the second scenario, where we do not know all the PDE terms, but have a prior knowledge of all possible candidate terms, several approaches on PDE discovery have also been developed~\cite{brunton2016discovering, rudy2017data}. In a general setup, discovering PDEs only from data without any prior knowledge is much more difficult~\cite{chen2022symbolic, du2022discover}. To address this challenge, instead of discovering the PDE in an explicit form, we use a neural network as an implicit representation of the physics laws. 


Our one-shot learning method first leverages the locality of PDEs and uses a neural network to learn the local solution operator of the PDE system (i.e., local PDE constraints) defined at a small computational domain. Then for a new PDE condition (e.g., a new source/forcing term or PDE coefficient field), we find the global PDE solution by coupling all local domains using the following mesh-based or neural network approaches. Specifically, a mesh-based fixed-point iteration (FPI) approach is proposed to obtain the PDE solution that satisfies the boundary/initial conditions and local PDE constraints. In this iterative approach, the computation on local stencil of mesh elements is in the same spirit as traditional PDE solvers. We also propose two versions of local-solution-operator informed neural networks (LOINNs), which are meshfree, to improve the stability and flexibility of finding the solution.
Moreover, our one-shot learning method has been applied to solve multi-dimensional, linear or nonlinear PDEs, and PDEs defined on a complex geometry. In this paper, we describe, in detail, the one-shot learning method for solution operators in Section \ref{sec: methods}, and then demonstrate on different PDEs for a range of conditions in Section \ref{sec: results}.
\section{Methods}
\label{sec: methods}

We first introduce the problem setup of learning solution operators of PDEs and then present our one-shot learning method.

\subsection{Learning solution operators of PDEs}
\label{sec:solution_operator}
We consider a physical system governed by a PDE defined on a spatio-temporal domain $\Omega \subset \mathbb{R}^d$:
\begin{equation*}
    \mathcal{F}[u(\mathbf{x}); f(\mathbf{x})] = 0, \quad \mathbf{x}=(x_1, x_2, \dots, x_d) \in \Omega
\end{equation*}
with suitable initial and boundary conditions 
\begin{equation}
\label{eq:bc}
    \mathcal{B} (u(\mathbf{x}), \mathbf{x}) = 0,
\end{equation}
where $u(\mathbf{x})$ is the solution of the PDE and $f(\mathbf{x})$ is a forcing term. The solution $u$ depends on $f$, and thus we define the solution operator as
$$\mathcal{G}: f(\mathbf{x}) \mapsto u(\mathbf{x}).$$
For nonlinear PDEs, $\mathcal{G}$ is a nonlinear operator.

In many problems, the PDE of a physical system is unknown or computationally expensive to solve, and instead, sparse data representing the physical system is available. Specifically, we consider a dataset $\mathcal{T}=\{(f_i,u_i)\}_{i=1}^{|\mathcal{T}|}$, and $(f_i,u_i)$ is the $i$-th data point, where $u_i=\mathcal{G}(f_i)$ is the PDE solution for $f_i$. Our goal is to learn $\mathcal{G}$ from the training dataset $\mathcal{T}$, such that for a new $f$, we can predict the corresponding solution $u=\mathcal{G}(f)$. When $\mathcal{T}$ is sufficiently large, then we can learn $\mathcal{G}$ straightforwardly by using neural networks, whose input and output are $f$ and $u$, respectively. Many networks have been proposed in this manner such as DeepONet~\cite{lu2021learning} and Fourier neural operator~\cite{li2020fourier}. In this study, we consider an extreme challenging scenario where we have only one data point for training, i.e., one-shot learning with $|\mathcal{T}|=1$, and we let $\mathcal{T}=\{(f_\mathcal{T}, u_\mathcal{T})\}$. 

\subsection{One-shot learning method based on the principle of locality}
\label{sec:one-shot}

It is impossible in general to learn the PDE solution operator $\mathcal{G}$ in Section~\ref{sec:solution_operator} from a single data point. To address this challenge, here we consider that $\mathcal{T}$ is not given, and we can select $f_\mathcal{T}$. In addition, instead of learning $\mathcal{G}$ for the entire input space, we only aim to predict $\mathcal{G}(f)$ in a neighborhood of some $f_0$. 

\begin{figure}[htbp]
    \centering
    \includegraphics[width = 14cm]{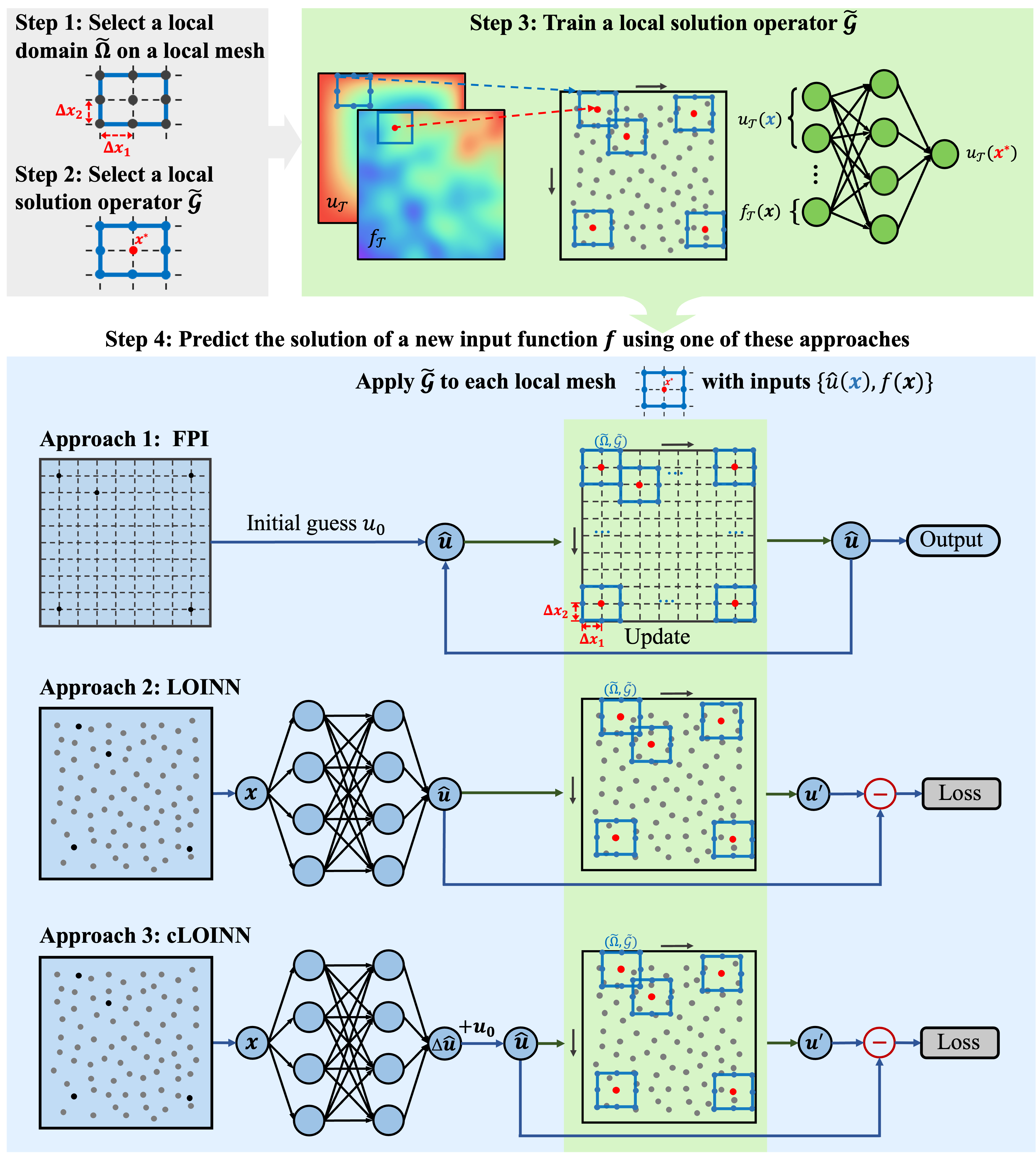}
    \caption{\textbf{Workflow of the one-shot learning method for solution operators.} (\textbf{Step 1}) We select a suitable polygon, such as a rectangle, on a local mesh with step size $\Delta x_1$ and $\Delta x_2$, and thus define a local domain $\tilde{\Omega}$ (the black nodes). (\textbf{Step 2}) We select a target mesh node $\mathbf{x}^*$ and define a local solution operator $\tilde{\mathcal{G}}$. (\textbf{Step 3}) We learn $\tilde{\mathcal{G}}$ using a neural network from a dataset constructed from $\mathcal{T} = (f_\mathcal{T}, u_\mathcal{T})$. (\textbf{Step 4}) For a new PDE condition (i.e., a new input function $f$), we utilize the pre-trained $\tilde{\mathcal{G}}$ to find the corresponding PDE solution by using one of the following approaches. (\textbf{Approach 1, FPI}) We consider points on an equispaced global mesh. Starting with an initial guess $u_0(\mathbf{x})$, we apply $\tilde{\mathcal{G}}$ iteratively to update the PDE solution until it is converged. (\textbf{Approach 2, LOINN}) We use a network to approximate the PDE solution. We apply $\tilde{\mathcal{G}}$ at different random locations to compute the loss function. (\textbf{Approach 3, cLOINN}) We use a network to approximate the difference between the PDE solution and the given $u_0(\mathbf{x})$.}
    \label{fig:workflow}
\end{figure}

To overcome the difficulty of training a machine learning model based on only one data point, we leverage the principle of locality that (partial) derivatives and differential equations are mathematically defined locally, i.e., the same PDE is satisfied in an arbitrary-shaped small domain inside $\Omega$. Based on this fact, instead of considering the entire computational domain, we consider a ``canonical'' small local domain $\tilde{\Omega}$, and we define a local solution operator $\tilde{\mathcal{G}}$ at $\tilde{\Omega}$. In our method, we can place $\tilde{\Omega}$ at any specific location inside $\Omega$, and each placement/realization leads to one training point for $\tilde{\mathcal{G}}$. In this way, we could easily generate a large training dataset for $\tilde{\mathcal{G}}$. Therefore, we transform the one-shot learning of $\mathcal{G}$ into a classical learning of $\tilde{\mathcal{G}}$. Once $\tilde{\mathcal{G}}$ is well trained, we predict $\mathcal{G}(f)$ for a new $f$ by applying $\tilde{\mathcal{G}}$ as a constrain at arbitrary local domains of the PDE solution $u = \mathcal{G}(f)$. Specifically, our method includes the following four steps (Fig.~\ref{fig:workflow}).

\begin{enumerate}
    \item[Step 1] \textbf{Select a ``canonical'' local domain $\tilde{\Omega}$ (Section~\ref{sec:select_local}).} We consider a virtual background equispaced mesh and select a polygon on this local mesh. We denote the set of all the mesh nodes that lie on the edges and within the interior of the chosen polygon (marked as black nodes in Fig.~\ref{fig:workflow}) as $\tilde{\Omega}$.
    \item[Step 2] \textbf{Select a local solution operator $\tilde{\mathcal{G}}$ (Section~\ref{sec:select_local}).} We choose a specific location $\mathbf{x}^*$ (the red node in Fig.~\ref{fig:workflow}) from the local domain $\tilde{\Omega}$ in step 1, and then the other points $\tilde{\Omega}_{\text{aux}} = \{\mathbf{x} \in \tilde{\Omega} | \mathbf{x}\neq\mathbf{x}^*\}$ are called ``auxiliary points'' (the blue nodes in Fig.~\ref{fig:workflow}). We construct a local solution operator 
    $$
    \tilde{\mathcal{G}}: \left( \{u(\mathbf{x}): \mathbf{x} \in \tilde{\Omega}_{\text{aux}} \} , \{f(\mathbf{x}): \mathbf{x} \in \tilde{\Omega} \} \right) \mapsto u(\mathbf{x}^*).$$
    This local solution operator $\tilde{\mathcal{G}}$ is learned by a fully-connected neural network that takes $u$ values of auxiliary points and $f$ values in $\tilde{\Omega}$ as inputs and outputs $u$ value of $\mathbf{x}^*$, which captures the local relationship of the PDE. 
    \item[Step 3] \textbf{Train the local solution operator $\tilde{\mathcal{G}}$ (Section~\ref{sec:learn_local}).} Utilizing the dataset $\mathcal{T} = (f_\mathcal{T}, u_\mathcal{T})$, we place the local domain $\tilde{\Omega}$ at different locations of $\Omega$, which can be either a global structured mesh or randomly sampled locations. This process generates many input-output data pairs for training $\tilde{\mathcal{G}}$. 
    \item[Step 4] \textbf{Predict the solution $u$ for a new input function $f$ (Sections~\ref{sec:fpi} and \ref{sec:loinn}).} For a new PDE condition $f$, we choose one of the three approaches to find the global PDE solution using the pre-trained local solution operator $\tilde{\mathcal{G}}$: fixed-point iteration (FPI, Section~\ref{sec:fpi}), local-solution-operator informed neural network (LOINN) or local-solution-operator informed neural network with correction (cLOINN) in Section~\ref{sec:loinn}.
\end{enumerate}

FPI is a mesh-based approach, and we can only use the structured equispaced global mesh to predict the solution. In contrast, with LOINN and cLOINN, it is possible to train the neural networks using randomly sampled data points in the domain. When we use equispaced global grid in LOINN and cLOINN, we denote the methods as LOINN-grid and cLOINN-grid, respectively. When we use randomly sampled point locations, we denote the methods as LOINN-random and cLOINN-random. In the following sections, we delve into the details of each step in our one-shot learning method for PDE solution operators.

\subsubsection{Selecting a local domain and a local solution operator}
\label{sec:select_local}

\begin{figure}[htbp]
    \centering
    \includegraphics[width=13cm]{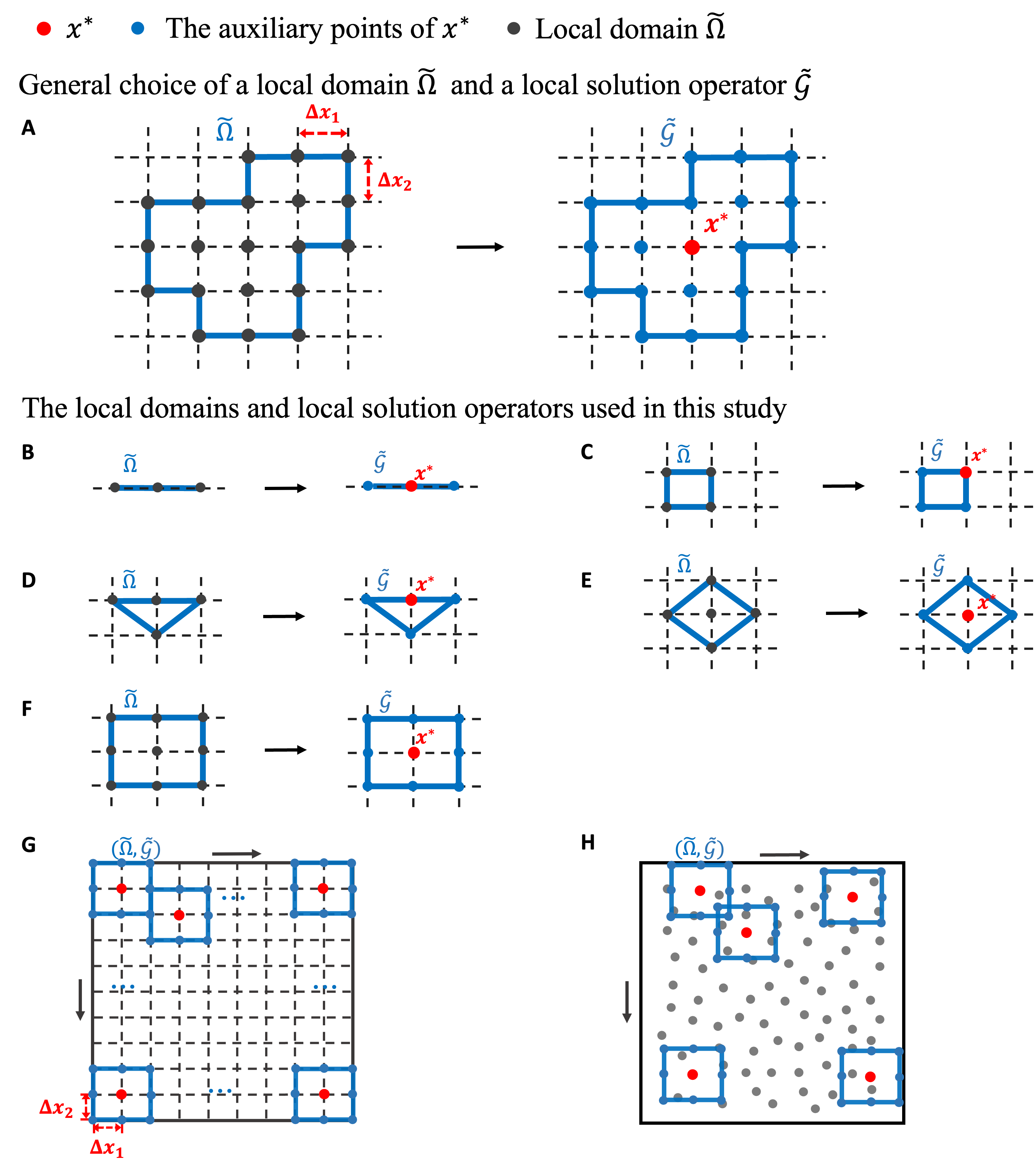}
    \caption{\textbf{Selecting local domains and learning local solution operators.} (\textbf{A}) A general choice of the local domain $\tilde{\Omega}$ is a set of nodes on a polygon. $\tilde{\Omega}$ can have different shapes and sizes according to a specific PDE. A local solution operator $\tilde{\mathcal{G}}$ is defined on $\tilde{\Omega}$. (\textbf{B} to \textbf{F}) Examples of local domains and local solution operators used in this paper. (\textbf{G} and \textbf{H}) When learning $\tilde{\mathcal{G}}$, the training points based on $\tilde{\Omega}$ are either (G) selected on a global mesh or (H) randomly sampled.}
    \label{fig:mesh}
\end{figure}

In this section, we elaborate on the first two steps of our method. To begin with, we consider a virtual background equispaced mesh. This is not a real mesh for the solution prediction in the computational domain; instead, it is a virtual mesh that guides us to select a local domain. Taking a two-dimensional problem as an example, this local equispaced mesh has mesh size $\Delta x_1$ and $\Delta x_2$, where $x_1$ and $x_2$ are spatial or temporal dimensions corresponding to the PDE. On this local mesh, we then sketch a polygon with mesh nodes positioned on its edges or within the interior of the polygon. We denote the set of these mesh nodes as $\tilde{\Omega}$, and we show a general choice of $\tilde{\Omega}$ in Fig.~\ref{fig:mesh}A (left) by black nodes.

In the second step, we choose a location $\mathbf{x}^*$ from the local domain $\tilde{\Omega}$ as the target node, and the remaining nodes surrounded $\mathbf{x}^*$ are denoted as auxiliary points $\tilde{\Omega}_{\text{aux}} = \{\mathbf{x} \in \tilde{\Omega} | \mathbf{x}\neq\mathbf{x}^*\}$. We define a local solution operator to predict the PDE solution at $\mathbf{x}^*$ from the information of auxiliary points and the PDE condition 
$$
\tilde{\mathcal{G}}: \left( \{u(\mathbf{x}): \mathbf{x} \in \tilde{\Omega}_{\text{aux}} \} , \{f(\mathbf{x}): \mathbf{x} \in \tilde{\Omega} \} \right) \mapsto u(\mathbf{x}^*),
$$
which is learned by a neural network. The intuition of this definition is that, for a well-defined PDE, if we know the solution $u$ at the boundary of $\tilde{\Omega}$ and $f$ within $\tilde{\Omega}$, then the solution $u$ at $\mathbf{x}^*$ inside $\tilde{\Omega}$ is determined. We find that only using the value of $f$ at $\mathbf{x}^*$ is sufficient, so in this study the local solution operator is chosen as
$$
\tilde{\mathcal{G}}: \left( \{u(\mathbf{x}): \mathbf{x} \in \tilde{\Omega}_{\text{aux}} \} , f(\mathbf{x}^*) \right) \mapsto u(\mathbf{x}^*).
$$

The choice of the shape and size of $\tilde{\Omega}$ and $\tilde{\mathcal{G}}$ is problem dependent. In Figs.~\ref{fig:mesh}B--F, we present several choices of $\tilde{\Omega}$ and $\tilde{\mathcal{G}}$ used in this paper.  

\subsubsection{Learning a local solution operator}
\label{sec:learn_local}

To train the local solution operator $\tilde{\mathcal{G}}$, we construct a training dataset from $\mathcal{T}=\{(f_\mathcal{T}, u_\mathcal{T})\}$. To make the network generalizable to different $f$, the selection of $f_\mathcal{T}$ in $\mathcal{T}$ is important. In our study, we use $f_{\mathcal{T}}(x) = f_{\text{random}}(x) + f_0(x)$, where $f_{\text{random}}(x) \sim \mathcal{GP}(0, k(x_1, x_2))$ is randomly sampled from a mean-zero Gaussian random field (GRF) with the covariance kernel $k(x_1,x_2)=\sigma^2 \exp(-\|x_1-x_2\|^2/2l^2)$ ($\sigma$: amplitude; $l$: length scale). The randomness in $f_{\text{random}}(x)$ is used to induce a more ``diverse'' training dataset so that we can extract more information from it. 

Since $\tilde{\mathcal{G}}$ learns for the small local domain $\tilde{\Omega}$, to generate many input-output pairs for training $\tilde{\mathcal{G}}$, we have two approaches. In the first approach, we have a structured equispaced global mesh $\hat{\Omega}$ with mesh size $\Delta x_1$ and $\Delta x_2$ (Fig.~\ref{fig:mesh}G). Note that the mesh size of the global mesh matches the local mesh. We place $\tilde{\Omega}$ at different locations of the global mesh $\hat{\Omega}$, and each placement would lead to one input-output data pair. In the second approach, we randomly sample different locations in $\Omega$ (Fig.~\ref{fig:mesh}H). Similarly, each location leads to one data point. These approaches make it possible to learn a network from only one PDE solution by converting one-shot learning to classical learning. For example, for a PDE in a spatio-temporal domain, we may choose the local domain Fig.~\ref{fig:mesh}C and generate a dataset 
$$
\mathcal{D}_\mathcal{T} = \Bigl\{\underbrace{\Bigl[u(x, t-\Delta t), u(x-\Delta x, t), u(x + \Delta x, t), f(x, t)\Bigl]}_{\text{Input}}, \underbrace{u(x, t)}_{\text{Output}}\Bigl\}_{(x,t)},
$$ 
where $(x,t)$ is sampled from the aforementioned approaches. We discuss the reasoning of using neural networks to learn the local solution operators in Appendix~\ref{sec:nn_reason}. 

Since the inputs of the neural network $\Tilde{\mathcal{G}}$ also include the solution to be predicted, we cannot predict solution for a new $f = f_0 + \Delta f$ directly. To address this issue, we propose three approaches to predict the solution of a new $f$ in Sections~\ref{sec:fpi} and \ref{sec:loinn}, whose computational efficiency and prediction accuracy depend on the choice of $\tilde{\mathcal{G}}$.



\subsubsection{Prediction via a fixed-point iteration (FPI)}
\label{sec:fpi}
In the step 4 of the one-shot learning method, we first propose a mesh-based fixed-point iteration approach (Algorithm~\ref{alg:FPI_algo}; Fig.~\ref{fig:workflow}, Approach 1). The solution is only considered on a equispaced global mesh $\hat{\Omega}$ with the same mesh size $\Delta x_1$ and $\Delta x_2$ that matches the local mesh $\tilde{\Omega}$. Because $f$ is close to $f_0$, we use $u_0 = \mathcal{G}(f_0)$ as the initial guess of $u$, and then in each iteration, we apply the pre-trained $\tilde{\mathcal{G}}$ at the current solution as the input to get a updated solution. When the solution is converged, $u$ and $f$ are consistent with respect to the local operator $\tilde{\mathcal{G}}$, and thus the current $u$ is the solution of our PDE.

\begin{algorithm}[htbp]
\caption{\textbf{Predict the solution $u=\mathcal{G}(f)$ for a new $f$ via FPI.}}
\label{alg:FPI_algo}
\KwIn{\text{Pre-trained model} $\tilde{\mathcal{G}}$\text{, an initial guess} $u_{0} = \mathcal{G}(f_{0})$\text{, a global mesh }$\hat{\Omega}$}
Initiate: $u(\mathbf{x}) \leftarrow u_{0}(\mathbf{x})$ for all $\mathbf{x} \in \hat{\Omega}$\;
\While{$u$ has not converged}{
  \For{$\mathbf{x} \in \hat{\Omega}$}{
        Construct $\tilde{\Omega}_{\text{aux, }\mathbf{x}}$ for the location $\mathbf{x}$\;
        $\hat{u}(\mathbf{x}) \leftarrow \tilde{\mathcal{G}} \left( \{u(\xi): \xi\in \tilde{\Omega}_{\text{aux, }\mathbf{x}} \} , f(\mathbf{x}) \right)$\;
}
Apply boundary and initial conditions to $\hat{u}(\mathbf{x})$\;
Update: $u(\mathbf{x}) \leftarrow \hat{u}(\mathbf{x})$ for all $\mathbf{x} \in \hat{\Omega}$\;
}
\KwOut{\text{$u(\mathbf{x})$ for all $\mathbf{x} \in \hat{\Omega}$} }
\end{algorithm}

\subsubsection{Prediction via a local-solution-operator informed neural network}
\label{sec:loinn}

We also propose a neural network-based approach LOINN (Fig.~\ref{fig:workflow}, Approach 2), which is meshfree and has more flexibility to handle boundary/initial conditions and training points inside the computational domain. We construct a neural network with parameters $\theta$ that takes the coordinates $\mathbf{x}$ as the input, and output the approximated solution $\hat{u}(\mathbf{x}; \theta)$. To train the network, we define the loss function that constrains $\hat{u}$ to satisfy $\Tilde{\mathcal{G}}$ at some local domains:
\begin{equation}
\label{eq:loinnpde}
\mathcal{L}_{\text{PDE}}(\theta) = \frac{1}{|\mathcal{T}_l|} \sum_{\mathbf{x} \in \mathcal{T}_l} \left(\hat{u}(\mathbf{x}; \theta) - \tilde{\mathcal{G}} \left( \{\hat{u}(\xi): \xi\in \tilde{\Omega}_{\text{aux, }\mathbf{x}} \} , f(\mathbf{x}) \right) \right)^2,
\end{equation}
where $\mathcal{T}_l$ is a set of point locations in the domain. The points can be sampled on a global mesh or randomly sampled (Hammersly sequence used in this study). For the boundary and initial condition in Eq.~\eqref{eq:bc}, similar to physics-informed neural networks~\cite{raissi2019physics,lu2021deepxde}, we define another loss function
\begin{equation}
\label{eq:lloinn}
\mathcal{L}_{\text{IC/BC}}(\theta) = \frac{1}{|\mathcal{T}_b|} \sum_{\mathbf{x} \in \mathcal{T}_b} \lVert \mathcal{B} (\hat{u}(\mathbf{x}; \theta), \mathbf{x}) \lVert _2^2, 
\end{equation}
where $\mathcal{T}_b$ is a set of point locations on the boundary or initial domain. Then the total loss is 
\begin{equation}
\mathcal{L}(\theta) = \mathcal{L}_{\text{PDE}}(\theta) + \mathcal{L}_{\text{IC/BC}}(\theta).
\end{equation}
In some cases, the initial and boundary conditions can be directly imposed by modifying the network architecture~\cite{lu2021physics}, which eliminates the necessity of the loss $\mathcal{L}_{\text{IC/BC}}$.


To improve the performance of LOINN, we develop LOINN with correction (cLOINN). Specifically, we modify the network architecture by using a last layer of adding $u_0$ (Fig.~\ref{fig:workflow}, Approach 3), and then the solution is $\hat{u}(\mathbf{x}) = \mathcal{N}(\mathbf{x}) + u_0(\mathbf{x})$, where $\mathcal{N}(\mathbf{x})$ is the original neural network output.

\section{Results}
\label{sec: results}

In this section, we demonstrate the capability and potential of our proposed one-shot learning method through various problems, and discuss choices of the local solution operator $\Tilde{\mathcal{G}}$, variations in $\Delta f$, and different mesh resolutions. Our numerical experiments cover a range of representative scenarios, including multi-dimensional, linear and nonlinear PDE systems, PDEs with $f$ either as a source term or a coefficient field, and PDEs defined in a complex geometry (Table~\ref{tab:L2error_more}). 

In each experiment, we first obtain the training dataset via finite difference methods on an equispaced dense mesh. The parameters for generating datasets and the hyperparameters of pre-trained neural networks are listed in Appendix~\ref{sec:hyperprarmeters}.
To test the developed method, we randomly sample 100 new $f$ by $f = f_0 + \Delta f$, in which $\Delta f$ is sampled from a GRF with a correlation length $l = 0.1$ and various amplitude $\sigma_{\text{test}}$. We compute the geometric mean and standard deviation of the $L^2$ relative errors for the 100 test cases. For all experiments, the Python library DeepXDE \cite{lu2021deepxde} is utilized to implement the neural networks. 


\begin{table}[htbp]
    \caption{\textbf{$L^2$ relative errors of one-shot learning method.} $\sigma_{\text{test}}$ is the amplitude of the GRF from which the test $\Delta f$ is sampled. LOINN-grid and LOINN-random represent LOINN approach using grid points and randomly sampled points, respectively. The same for cLOINN. Since cLOINN-random performs the best among LOINN/cLOINN methods, we only show FPI and cLOINN-random for some examples.}
    \footnotesize
    \label{tab:L2error_more}
    \centering
    \begin{tabular}{lcccccc}
    \toprule
    & $\sigma_{\text{test}}$ & FPI & LOINN-grid & cLOINN-grid & LOINN-random & cLOINN-random\\
    \midrule
    & 0.02 & 0.98 $\pm$ 0.49\% & 0.98 $\pm$ 0.49\% & 1.02 $\pm$ 0.47\%  & 0.98 $\pm$ 0.49\% & 0.98 $\pm$ 0.58\% \\
    Section~\ref{sec: Poisson1D} & 0.05 & 1.67 $\pm$ 1.28\% & 1.67 $\pm$ 1.28\% & 1.71 $\pm$ 1.28\% & 1.67 $\pm$ 1.28\% & 1.88 $\pm$ 1.64\% \\
    1D Poisson
    & 0.10 & 3.71 $\pm$ 2.96\% & 3.71 $\pm$ 2.96\% & 3.82 $\pm$ 3.12\% & 3.71 $\pm$ 2.96\% & 4.33 $\pm$ 3.30\% \\
    & 0.15 & 5.55 $\pm$ 5.17\% & 5.55 $\pm$ 5.17\% & 5.69 $\pm$ 5.19\% & 5.54 $\pm$ 5.18\% & 6.33 $\pm$ 5.57\% \\
    \midrule
    & 0.10 & 0.45 $\pm$ 0.11\% & 2.90 $\pm$ 0.85\% & 0.92 $\pm$ 0.40\%  & 0.82 $\pm$ 0.23\% & 0.62 $\pm$ 0.11\%\\
    Section~\ref{sec: LinearDiffusion}
    & 0.30 & 1.26 $\pm$ 0.43\% & 3.36 $\pm$ 1.03\% & 1.61 $\pm$ 0.50\% & 1.85 $\pm$ 0.56\% & 1.33 $\pm$ 0.46\%\\
    Linear diffusion
    & 0.50 & 2.66 $\pm$ 1.48\% & 4.58 $\pm$ 1.60\% & 2.89 $\pm$ 1.64\% & 3.59 $\pm$ 2.42\% & 2.70 $\pm$ 1.46\%\\
    & 0.80 & 5.22 $\pm$ 3.63\% & 7.15 $\pm$ 3.34\% & 5.59 $\pm$ 4.09\% & 5.99 $\pm$ 3.57\% & 5.21 $\pm$ 3.61\% \\
    \midrule
    & 0.10 & 0.30 $\pm$ 0.06\% & 0.46 $\pm$ 0.14\% & 0.54 $\pm$ 0.21\%  & 0.32 $\pm$ 0.07\% & 0.34 $\pm$ 0.11\%\\
    Section~\ref{sec: NonlinearDiffusion}
    & 0.30 & 0.78 $\pm$ 0.23\% & 0.99 $\pm$ 0.30\% & 0.94 $\pm$ 0.25\% & 0.81 $\pm$ 0.23\% & 0.99 $\pm$ 0.49\%\\
    Nonlinear diffusion-reaction
    & 0.50 & 1.39 $\pm$ 0.49\% & 1.57 $\pm$ 0.55\% & 1.46 $\pm$ 0.46\% & 1.41 $\pm$ 0.49\% & 1.61 $\pm$ 0.87\%\\
    & 0.80 & 2.32 $\pm$ 1.32\% & 2.45 $\pm$ 1.31\% & 2.38 $\pm$ 1.30\% & 2.34 $\pm$ 1.31\% & 3.24 $\pm$ 2.33\% \\
    \bottomrule
\end{tabular}

\begin{tabular}{lcccc}
    \toprule
    && $\sigma_{\text{test}}$ & FPI & cLOINN-random\\
    \midrule
    & \multirow{2}{*}{}
    & 0.50 & 0.81 $\pm$ 0.32\% & 0.96 $\pm$ 0.33\%  \\
    Section~\ref{sec: Advection} && 1.00 & 1.91 $\pm$ 1.88\% & 2.14 $\pm$ 1.80\% \\
    Advection
    && 1.50 & 4.32 $\pm$ 4.31\% & 4.57 $\pm$ 4.26\% \\
    && 2.00 & 8.25 $\pm$ 9.47\% & 8.70 $\pm$ 9.32\% \\
    \midrule 
    & \multirow{4}{*}{$\Tilde{\mathcal{G}}_1$} 
    & 0.05 & 2.58 $\pm$ 0.63\% & 4.82 $\pm$ 1.60\%  \\
    && 0.10 & 3.62 $\pm$ 1.35\% & 9.43 $\pm$ 3.69\% \\
    && 0.20 & 5.88 $\pm$ 2.97\% & 17.71 $\pm$ 5.97\% \\
    Section~\ref{sec: Poisson2D} && 0.30 & 9.11 $\pm$ 4.03\% & 27.16 $\pm$ 10.40\% \\
    \cmidrule{2-5}
    2D nonlinear Poisson
    & \multirow{4}{*}{$\Tilde{\mathcal{G}}_2$} 
    & 0.05 & 1.68 $\pm$ 0.81\% & 4.29 $\pm$ 1.75\% \\
    && 0.10 & 2.75 $\pm$ 1.32\% & 9.16 $\pm$ 3.82\%\\
    && 0.20 & 4.49 $\pm$ 2.43\% & 17.69 $\pm$ 5.94\%\\
    && 0.30 & 8.02 $\pm$ 4.52\% & 28.06 $\pm$ 10.60\% \\
    \midrule
    && 0.05 & 2.20 $\pm$ 0.93\% & 2.14 $\pm$ 0.95\%  \\
    Section~\ref{sec: Poisson2D_cutout} && 0.10 & 3.21 $\pm$ 1.82\% & 2.88 $\pm$ 1.85\% \\
    2D nonlinear Poisson with a circle cutout && 0.20 & 6.59 $\pm$ 3.55\% & 6.23 $\pm$ 3.66\% \\
    && 0.30 & 14.11 $\pm$ 7.62\% & 13.86 $\pm$ 7.93\% \\
    \midrule
    & \multirow{4}{*}{$C_A$} 
    & 0.10 & 0.57 $\pm$ 0.23\% & 3.31 $\pm$ 1.28\% \\
    && 0.30 & 1.50 $\pm$ 0.71\% & 9.20 $\pm$ 3.07\% \\
    && 0.50 & 2.14 $\pm$ 1.40\% & 13.56 $\pm$ 4.74\% \\
    Section~\ref{sec: PorousMedia} && 0.80 & 4.07 $\pm$ 3.36\% & 16.01 $\pm$ 3.78\% \\  
    \cmidrule{2-5}
    Diffusion-reaction system in porous media
    & \multirow{4}{*}{$C_B$} 
    & 0.10 & 0.54 $\pm$ 0.17\% & 3.14 $\pm$ 1.14\%  \\
    && 0.30 & 1.49 $\pm$ 0.91\% & 9.13 $\pm$ 4.29\% \\
    && 0.50 & 2.48 $\pm$ 2.41\% & 15.13 $\pm$ 9.80\% \\
    && 0.80 & 4.68 $\pm$ 4.22\% & 24.08 $\pm$ 17.22\% \\
    \bottomrule
\end{tabular}

\end{table}

\subsection{1D Poisson equation}
\label{sec: Poisson1D}

We first demonstrate the capability of our method with a pedagogical example of a one-dimensional Poisson equation
\[\Delta u = 100f(x), \quad x \in [0, 1]\]
with the zero Dirichlet boundary condition, and the solution operator is $\mathcal{G}: f\mapsto u$. 

\begin{figure}[htbp]
    \centering    \includegraphics[width=\textwidth]{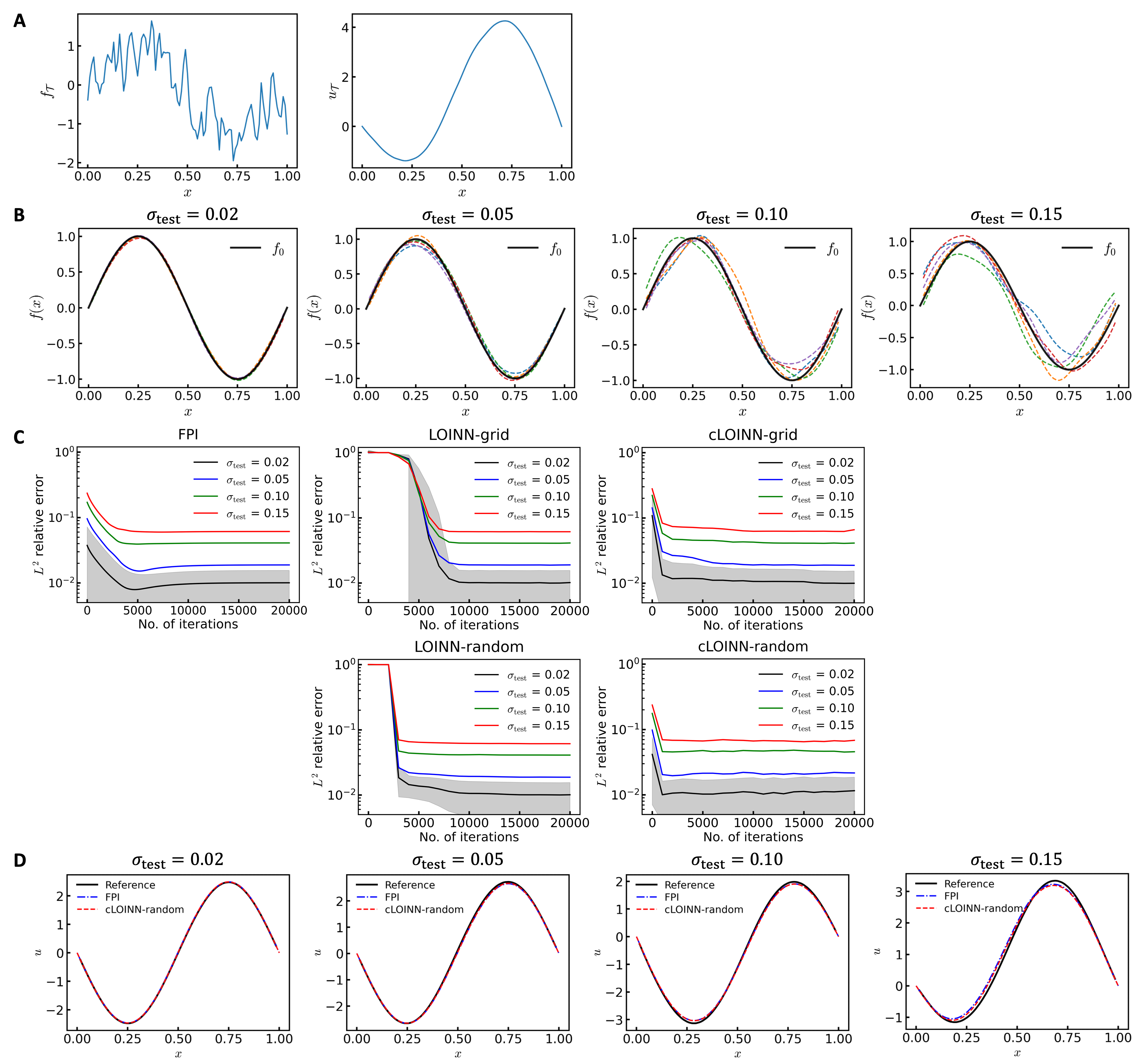}
    \caption{\textbf{1D Poisson equation in Section~\ref{sec: Poisson1D}.}
    (\textbf{A}) The training data includes a random $f_{\mathcal{T}}$ generated from GRF and the corresponding solution $u_{\mathcal{T}}$. 
    (\textbf{B}) Testing examples of random $f = f_0 + \Delta f$ with $\Delta f$ sampled from GRF of $\sigma_{\text{test}} = 0.02, 0.05, 0.10, 0.15$ and $l = 0.1$.
    (\textbf{C}) The convergence of $L^2$ relative errors of different approaches for various test cases.
    (\textbf{D}) Prediction example of different approaches for various test cases.}   
    \label{fig:Poisson1D_res}
\end{figure}

We choose the simplest local solution operator using 3 nodes  (Fig.~\ref{fig:mesh}B)
$$\Tilde{\mathcal{G}}: \{u(x - \Delta x), u(x + \Delta x),f(x)\} \mapsto u(x)$$
with $\Delta x = 0.01$. The training dataset $\mathcal{T}$ only has one data point $(f_\mathcal{T}, u_\mathcal{T})$ (Fig.~\ref{fig:Poisson1D_res}A), where $f_\mathcal{T}$ is generated based on $f_0 = \sin(2\pi x)$. In Fig.~\ref{fig:Poisson1D_res}B, we show examples of testing cases $f = f_0 + \Delta f$ with different $\sigma_{\text{test}}$. When $\sigma_{\text{test}}$ is larger, there is a more significant discrepancy between $f_0$ and $f$. 

We evaluate the performance of FPI, LOINN and cLOINN approaches on the grid data, as well as LOINN and cLOINN using randomly sampled data points (Fig.~\ref{fig:Poisson1D_res}C). For FPI, LOINN-grid, and cLOINN-grid, we use a mesh with 101 equispaced points. For LOINN-random and cLOINN-random, we use 101 random points. We report the geometric mean of $L^2$ relative errors of all cases in Table~\ref{tab:L2error_more} for different $\sigma_{\text{test}}$. As expected, the smaller $\sigma_{\text{test}}$ we use, the better the performance. When $\sigma_{\text{test}} = 0.02$, all approaches achieve an $L^2$ relative error of around $1\%$. In Fig.~\ref{fig:Poisson1D_res}D, we show prediction examples of using FPI and cLOINN-random approaches. For this simple case, the results of using different approaches and data sampling are similar. It is observed that in this experiment, cLOINN converges faster than LOINN and FPI.


\subsection{Linear diffusion equation}
\label{sec: LinearDiffusion}

We consider a linear diffusion equation \[\frac{\partial u} {\partial t} = D \frac{\partial^2 u} {\partial x^2} + f(x), \quad x \in [0, 1], \quad t \in [0, 1]\]
with zero boundary and initial conditions, where $D=0.01$ is the diffusion coefficient. We aim to learn the solution operator $\mathcal{G}: f \mapsto u$ for a class of $f = f_0 + \Delta f$ with $f_0 = 0.9\sin(2\pi x)$.

\begin{figure}[htbp]
    \centering    \includegraphics[width=\textwidth]{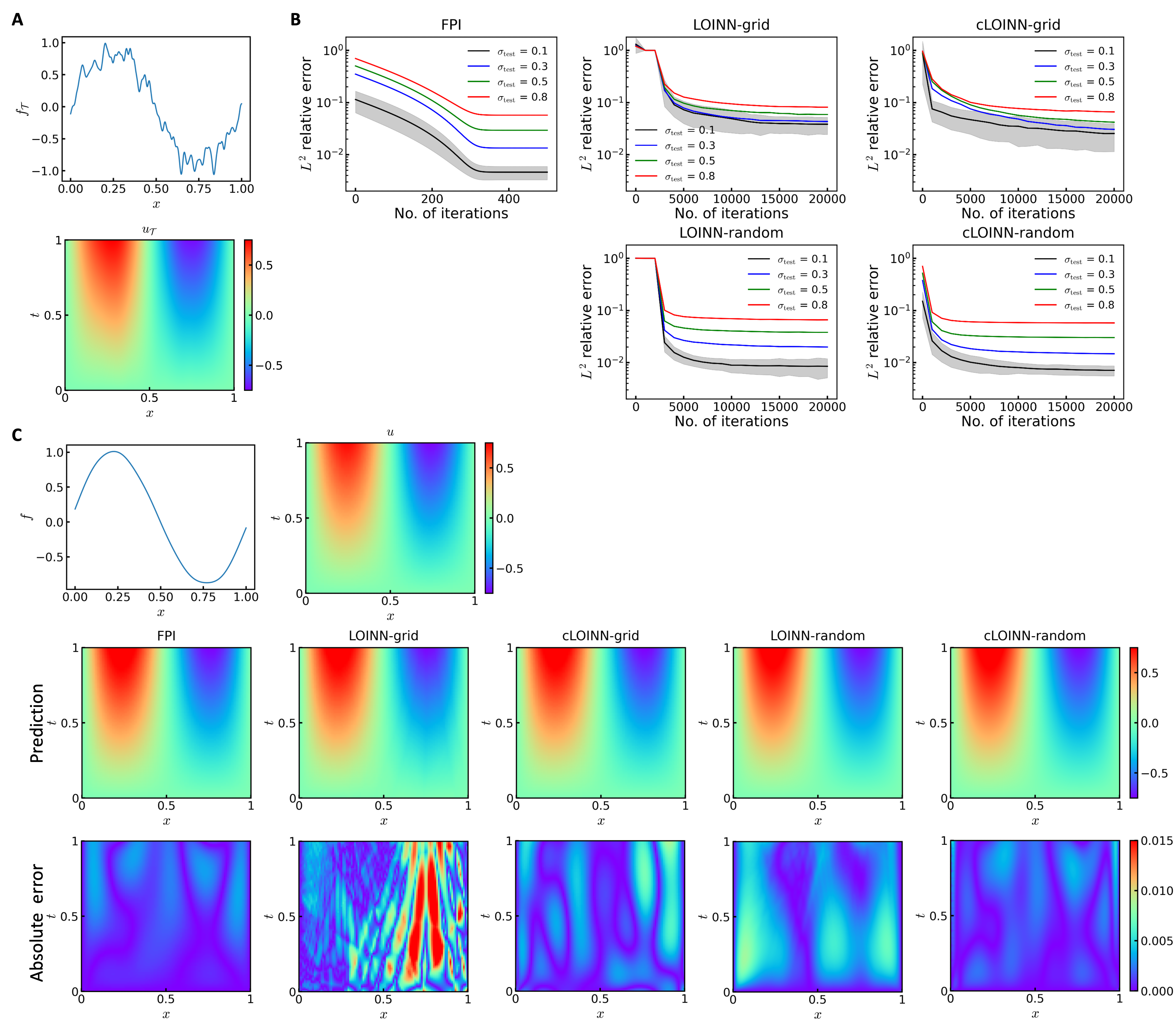}
    \caption{\textbf{Learning the linear diffusion equation in Section~\ref{sec: LinearDiffusion}.}
     (\textbf{A}) The training data includes a random $f_{\mathcal{T}}$ generated from GRF and the corresponding solution $u_{\mathcal{T}}$. 
     (\textbf{B}) The convergence of $L^2$ relative errors of different approaches for various test cases.
     (\textbf{C}) Prediction example of different approaches for a test case with $\sigma_{\text{test}} = 0.1$. }
    \label{fig:Diffusion_linear}
\end{figure}

We consider the simplest local solution operator defined on a local domain with 4 spatial-temporal nodes (Fig.~\ref{fig:mesh}D):
$$\Tilde{\mathcal{G}}: \{u(x, t-\Delta t), u(x-\Delta x , t), u(x+\Delta x, t), f(x, t)\} \mapsto u(x, t).$$ 
To generate the training dataset $\mathcal{T}$, we randomly sample $f_\mathcal{T}$ shown in Fig. \ref{fig:Diffusion_linear}A. FPI and LOINN/cLOINN-grid use an equispaced mesh of 101$\times$101, and LOINN/cLOINN-random use $101^2$ random point locations.

We test these approaches for $\Delta f$ sampled from GRF with $\sigma_{\text{test}} = $ 0.1, 0.3, 0.5, and 0.8. The details of error convergence are shown in Fig.~\ref{fig:Diffusion_linear}B. For a fixed $\sigma_{\text{test}}$, FPI and cLOINN-grid both work well and outperform LOINN-grid (Table~\ref{tab:L2error_more}). LOINN-random and cLOINN-random both achieve better accuracy than LOINN/cLOINN-grid (e.g., an $L^2$ relative error smaller than $1\%$ when $\sigma_{\text{test}} = 0.1$). 
When $\sigma_{\text{test}}$ is increased from 0.1 to 0.8, all approaches can achieve $L^2$ relative error smaller than $8\%$. 
In this experiment, we conclude that cLOINN performs better and converges faster than LOINN. Also, the performance of LOINN and cLOINN with randomly sampled points are better than that with mesh points. In Fig.~\ref{fig:Diffusion_linear}C, we show a test example for $\sigma_{\text{test}} = 0.1$ and the predictions and pointwise errors using FPI, LOINN-grid, cLOINN-grid, LOINN-random, and cLOINN-random. 

\subsection{Nonlinear diffusion-reaction equation}
\label{sec: NonlinearDiffusion}
We then consider a nonlinear diffusion-reaction equation with a source term $f(x)$:
\[\frac{\partial{u}}{\partial{t}} = D \frac{\partial^2 u}{\partial x^2} + k u^2 + f(x), \quad x \in [0, 1], t \in [0, 1]\]
with zero initial and boundary conditions, where $D=0.01$ is the diffusion coefficient, and $k=0.01$ is the reaction rate. The solution operator we aim to learn is $\mathcal{G}: f \mapsto u$, where $f = f_0 + \Delta f$ with $f_0 = \sin(2\pi x)$. 

\subsubsection{Learning the nonlinear diffusion-reaction equation}
\begin{figure}[htbp]
    \centering    \includegraphics[width=\textwidth]{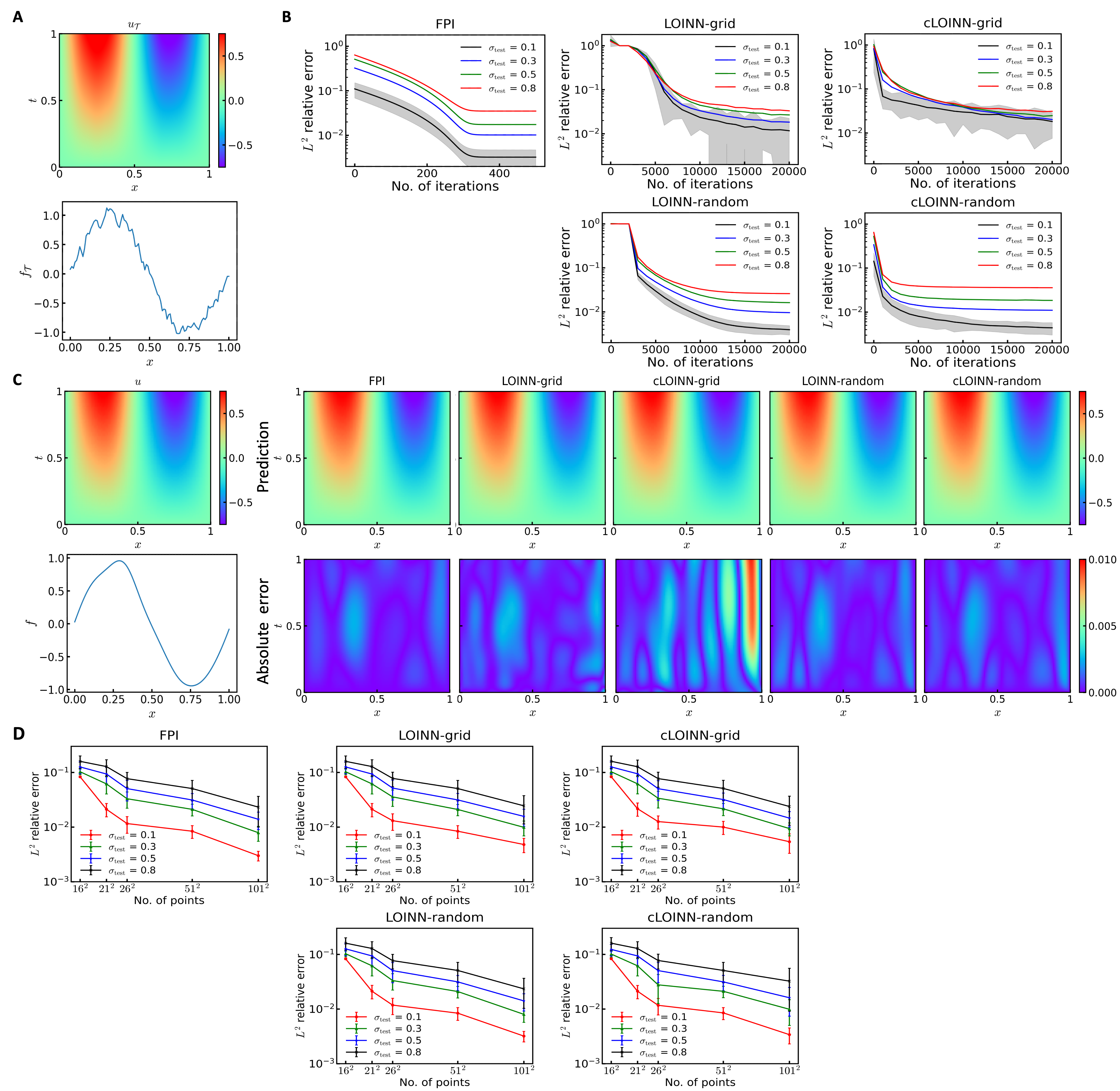}
    \caption{\textbf{Learning the nonlinear diffusion-reaction equation in Section~\ref{sec: NonlinearDiffusion}.}
     (\textbf{A}) The training data includes a random $f_{\mathcal{T}}$ generated from GRF and the corresponding solution $u_{\mathcal{T}}$. 
     (\textbf{B}) The convergence of $L^2$ relative errors of different approaches for various test cases.
     (\textbf{C}) Prediction example of different approaches for a test case with $\sigma_{\text{test}} = 0.1$. 
     (\textbf{D}) $L^2$ relative error of different test functions with $\sigma_{\text{test}} = 0.1, 0.3, 0.5, 0.8$ when using different number of point locations to show effect of mesh resolutions.}
     \label{fig:Diffusion_nonlinear}
\end{figure}

We use the same $\Tilde{\mathcal{G}}$ as the previous example in Section~\ref{sec: LinearDiffusion} (Fig.~\ref{fig:mesh}D), and randomly sample $f_\mathcal{T}$ shown in Fig. \ref{fig:Diffusion_nonlinear}A. FPI and LOINN/cLOINN-grid use an equispaced mesh of 101$\times$101, and LOINN/cLOINN-random use $101^2$ random point locations. We also test these approaches for $\Delta f$ sampled from GRF with $\sigma_{\text{test}} = $ 0.1, 0.3, 0.5, and 0.8 (Fig.~\ref{fig:Diffusion_nonlinear}B). FPI and LOINN/cLOINN-random achieve better accuracy than the others (e.g., $L^2$ relative error smaller than $0.5\%$ when $\sigma_{\text{test}} = 0.1$). 
When $\sigma_{\text{test}}$ is increased from 0.1 to 0.8, all the approaches can achieve $L^2$ relative error smaller than $5\%$. In Fig.~\ref{fig:Diffusion_nonlinear}C, we show a test example for $\sigma_{\text{test}} = 0.1$ and the predictions and pointwise errors using FPI, LOINN-grid, cLOINN-grid, LOINN-random, and cLOINN-random. 

\subsubsection{Analyzing the effect of mesh resolutions of local solution operator}

In all the previous examples, we use a local mesh with resolution $\Delta x = \Delta t = 0.01$ for learning the local solution operator $\tilde{\mathcal{G}}$. In this section, we investigate the performance of our methods using different mesh resolutions when training $\Tilde{\mathcal{G}}$ and predicting solutions in $\Omega$. For FPI, LOINN-grid, and cLOINN-grid, we generate input-output pairs for training $\tilde{\mathcal{G}}$ on a structured equispaced global mesh $\hat{\Omega}$ with mesh size $\Delta x = \Delta t = $ 0.01, 0.02, 0.04, 0.05, and 0.07 (i.e., the numbers of points are $101^2, 51^2, 26^2, 21^2$, and $16^2$), which matches the local mesh resolutions. For LOINN-random and cLOINN-random, we randomly sample $101^2, 51^2, 26^2, 21^2, 16^2$ point locations in $\Omega$.
We compare $L^2$ relative errors with different $\sigma_{\text{test}}$ and resolutions. With denser mesh resolutions or more number of points, all approaches perform better (Fig.~\ref{fig:Diffusion_nonlinear}D). Even using a coarse mesh of 0.05, our methods can still achieve errors around 10\% for all cases, which demonstrates the computational efficiency of our proposed methods. 


\subsection{Advection equation}
\label{sec: Advection}
We test our method on an advection equation
\[\frac{\partial s} {\partial t} + a(x) \frac{\partial s} {\partial x} = 0, \quad x \in [0, 1], t \in [0, 1]\]
with initial condition $s(x, 0) = x^2$ and boundary conditions $s(0, t) = \sin(\pi t)$, where $a(x)$ is the velocity coefficient. We learn the solution operator mapping from the coefficient field to the PDE solution: $\mathcal{G}: a \mapsto s$ for a class of $a = a_0 + 0.1\Delta f$ with $a_0 = 1$.

\begin{figure}[htbp]
    \centering
    \includegraphics[width=\textwidth]{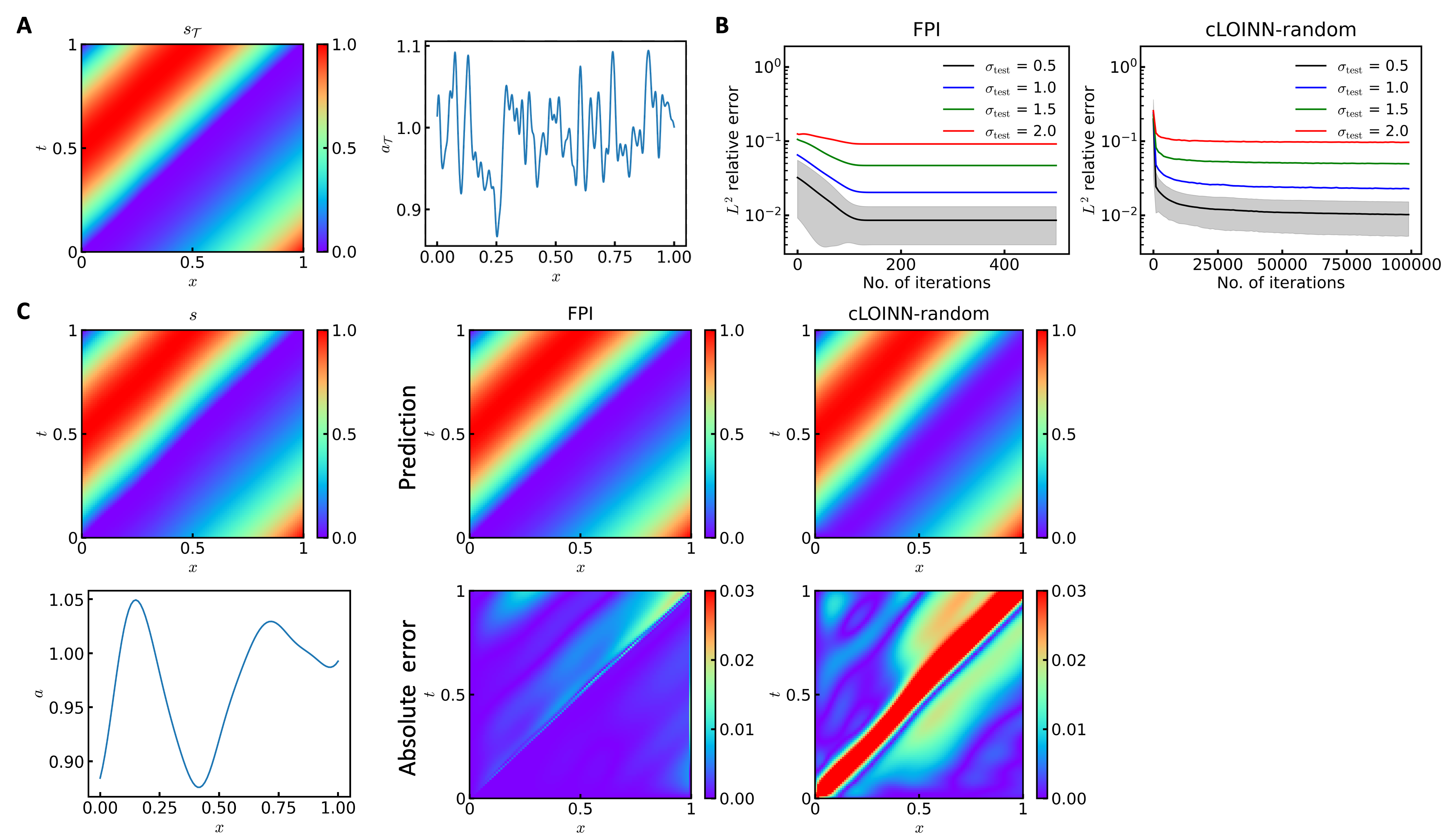}
    \caption{\textbf{Learning the advection equation in Section~\ref{sec: Advection}.} 
     (\textbf{A}) The training data includes a random $a_{\mathcal{T}}$ generated from GRF and the corresponding solution $s_{\mathcal{T}}$. 
     (\textbf{B}) The convergence of $L^2$ relative errors of FPI and cLOINN-random for various test cases.
     (\textbf{C}) Prediction example of FPI and cLOINN-random for a test case with $\sigma_{\text{test}} = 0.5$. }
    \label{fig:Advection_res}
\end{figure}

We use the local solution operator (Fig.~\ref{fig:mesh}C) 
$$\Tilde{\mathcal{G}}: \{s(x, t-\Delta t), s(x-\Delta x, t), s(x-\Delta x, t-\Delta t), a(x, t)\} \mapsto s(x, t).$$ 
The training dataset $\mathcal{T}$ with one data point $(a_\mathcal{T}, s_\mathcal{T})$ is shown in Fig.~\ref{fig:Advection_res}A. 
Since FPI and cLOINN-random have shown better performance compared to LOINN and cLOINN-grid in previous experiments, we only present the results of FPI and cLOINN-random here (Fig.~\ref{fig:Advection_res}B). FPI uses an equispaced mesh of 101$\times$101, and cLOINN-random use $101^2$ random point locations. The errors of FPI and cLOINN-random both achieve less than 1\% with $\sigma_{\text{test}} = 0.50$ (Table~\ref{tab:L2error_more}), with FPI outperforming cLOINN-random slightly. When $\sigma_{\text{test}}$ is increased from 0.50 to 2.00, the $L^2$ relative errors are smaller than $10\%$. We show a test example for $\sigma_{\text{test}} = 0.50$ and the prediction using FPI and cLOINN-random approaches in Fig.~\ref{fig:Advection_res}C.

\subsection{2D nonlinear Poisson equation in a square domain}
\label{sec: Poisson2D}

Next, we consider a 2D nonlinear Poisson equation 
\[\nabla((1+u^2)\nabla u) = 10f(x,y), \quad x \in [0, 1], y \in [0, 1]\]
with zero Dirichlet boundary conditions.  We aim to learn the solution operator $\mathcal{G}: f \mapsto u$ for a class of $f = f_0 + \Delta f$ with $f_0(x,y) = x\sin(y)$. 

In this example, we investigate the effect of the selection of the local domains. We choose two different local domains, including
a simple local domain with 5 nodes (Fig.~\ref{fig:mesh}E) $$\Tilde{\mathcal{G}}_1: \{u(x, y-\Delta y), u(x-\Delta x, y), u(x+\Delta x, y), u(x, y+\Delta y), f(x, y)\} \mapsto u(x, y) $$ 
and a larger domain with 9 nodes (Fig.~\ref{fig:mesh}F)
\[
\Tilde{\mathcal{G}}_2: \{u(x, y-\Delta y), u(x-\Delta x, y), u(x+\Delta x, y), u(x, y+\Delta y), u(x-\Delta x, y-\Delta y), u(x-\Delta x, y+\Delta y), \]
\[
u(x+\Delta x, y-\Delta y), u(x+\Delta x, y+\Delta y), f(x, y)\} \mapsto u(x, y).
\]
For this example, the numerical solution is obtained via the finite element method. 

\begin{figure}[htbp]
    \centering
    \includegraphics[width=\textwidth]{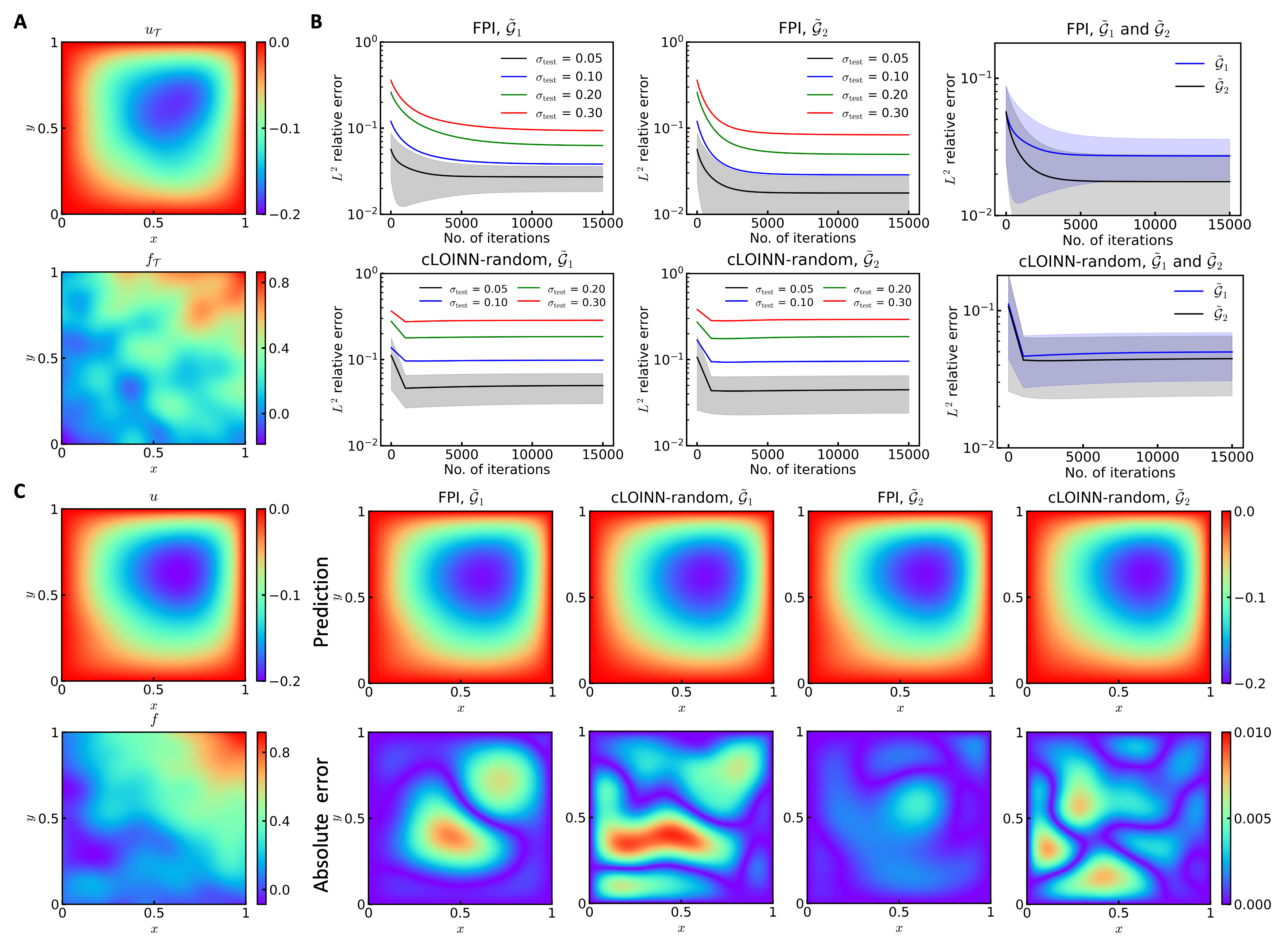}
    \caption{\textbf{Learning the 2D nonlinear Poisson equation in Section~\ref{sec: Poisson2D}.} 
     (\textbf{A}) The training data includes a random $f_{\mathcal{T}}$ generated from GRF and the corresponding solution $u_{\mathcal{T}}$. 
     (\textbf{B}) The convergence of $L^2$ relative errors of FPI and cLOINN-random for various test cases using $\Tilde{\mathcal{G}_1}$ and $\Tilde{\mathcal{G}_2}$.
     (\textbf{C}) Prediction example of different approaches for a test case with $\sigma_{\text{test}} = 0.05$.
    }
    \label{fig:Poisson2D_res}
\end{figure}

The training dataset $\mathcal{T}$ with one data point $(f_\mathcal{T}, u_\mathcal{T})$ is shown in Fig.~\ref{fig:Poisson2D_res}A. We compare the results of FPI and cLOINN-random using $\Tilde{\mathcal{G}}_1$ and $\Tilde{\mathcal{G}}_2$ (Table.~\ref{tab:L2error_more}). FPI performs well, and the $L^2$ relative errors achieve less than 2\% when $\sigma_{\text{test}} = 0.05$. Compared to FPI, the performance of cLOINN-random is worse but still acceptable when $\sigma_{\text{test}}$ is small, and the errors achieve less than 5\% when $\sigma_{\text{test}} = 0.05$. (Fig.~\ref{fig:Poisson2D_res}B). For both FPI and cLOINN-random, the local solution operator $\Tilde{\mathcal{G}_2}$ outperforms $\Tilde{\mathcal{G}_1}$ for $\sigma_{\text{test}} = 0.05, 0.10,$ and $0.20$. This makes sense since the size $\Tilde{\mathcal{G}_2}$ is larger. An example of the prediction using FPI and cLOINN-random approaches with $\Tilde{\mathcal{G}_1}$ is shown in Fig.~\ref{fig:Poisson2D_res}C.


\subsection{2D nonlinear Poisson equation in a square domain with a circle cutout}
\label{sec: Poisson2D_cutout}
\begin{figure}[htbp]
    \centering
    \includegraphics[width=\textwidth]{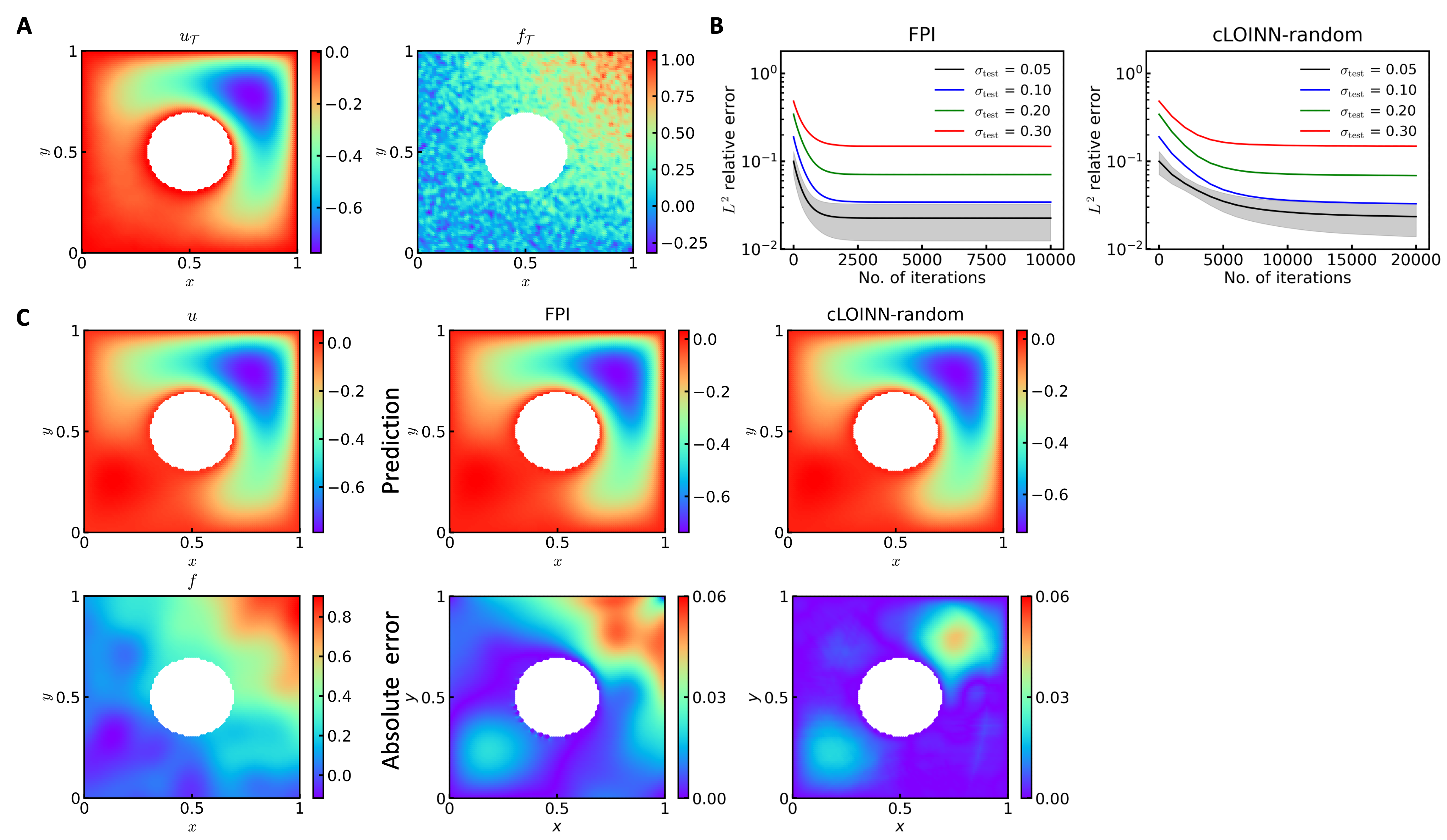}
    \caption{\textbf{Learning the 2D nonlinear Poisson equation with a circle cutout in Section~\ref{sec: Poisson2D_cutout}.} 
     (\textbf{A}) The training data includes a random $f_{\mathcal{T}}$ generated from GRF and the corresponding solution $u_{\mathcal{T}}$. 
     (\textbf{B}) The convergence of $L^2$ relative errors of FPI and cLOINN-random for various test cases.
     (\textbf{C}) Prediction example of different approaches for a test case with $\sigma_{\text{test}} = 0.05$.
    }
    \label{fig:Poisson2Dcutout_res}
\end{figure}

We also consider a 2D nonlinear Poisson equation in a square domain with a circle cutout of radius 0.2 (Fig.~\ref{fig:Poisson2Dcutout_res})
\[\nabla((1+u^2)\nabla u) = 100f(x,y)\]
with zero Dirichlet boundary conditions. We learn the solution operator $\mathcal{G}: f \mapsto u$ or a class of $f = f_0 + \Delta f$ with $f_0(x,y) = x\sin(y)$.

We choose a simple local domain $$\Tilde{\mathcal{G}}: 
\{u(x, y-\Delta y), u(x-\Delta x, y), u(x+\Delta x, y), u(x, y+\Delta y), f(x, y)\} \mapsto u(x, y) $$
with 5 nodes (Fig.~\ref{fig:mesh}E). For this example, the numerical solution is obtained using the finite element method. The domain is discretized using triangular elements, and the maximum size is 0.005. 

The training dataset $\mathcal{T}$ is shown in Fig.~\ref{fig:Poisson2Dcutout_res}A. When $\sigma_{\text{test}} = 0.05$, the $L^2$ relative errors achieve less than 3\%. Also cLOINN-random performs slightly better than FPI while FPI converges faster, as the values of cLOINN-random solution near the circle boundary are more accurate. The comparison between these two is shown in Fig.~\ref{fig:Poisson2Dcutout_res}B, and one example is shown in Fig.~\ref{fig:Poisson2Dcutout_res}C. We have shown that our approaches can work well in this complex geometry.

\subsection{Diffusion-reaction system in porous media}
\label{sec: PorousMedia}
We consider a diffusion-reaction system in porous media ($x \in [0, 1], t \in [0, 1]$)

\[\frac{\partial C_A} {\partial t}  = D \frac{\partial^2 C_A} {\partial x^2} - k_f C_A C_B^2 + f(x),\]
\[\frac{\partial C_B} {\partial t}  = D \frac{\partial^2 C_B} {\partial x^2} - 2k_f C_A C_B^2\]
with initial conditions $C_A(x, 0) = C_B(x,0) = e^{-20x}$ and zero Dirichlet boundary conditions, where $D=0.01$ is the diffusion coefficient, and $k_f=0.01$ is the reaction rate. The solution operator is $\mathcal{G}: f \mapsto (C_A, C_B)$. Here we predict the solutions $C_A$ and $C_B$ for a new $f = e^{-\frac{(x - 0.5)^2}{0.05}} + \Delta f$. 

\begin{figure}[htbp]
    \centering
    \includegraphics[width=16cm]{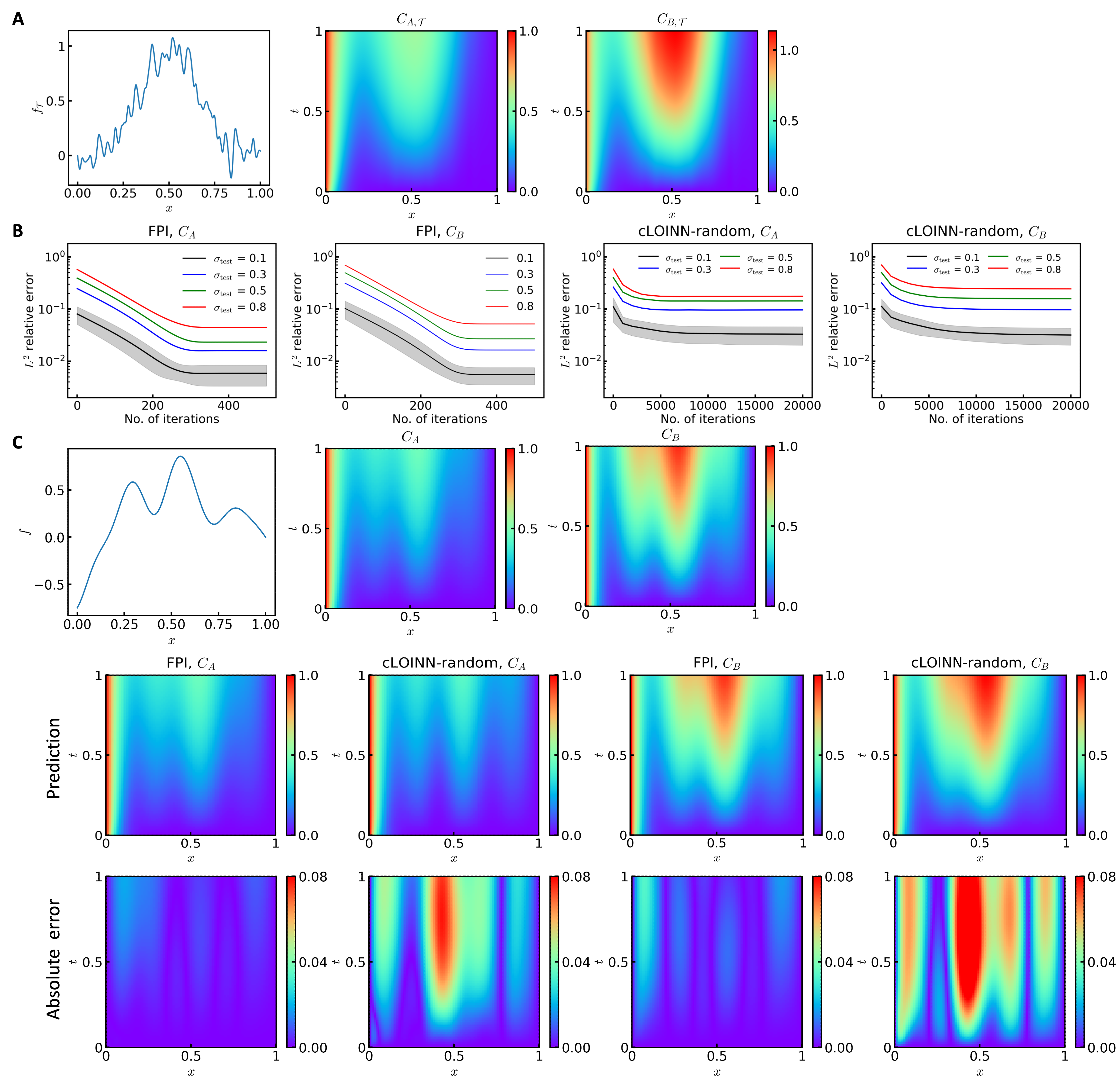}
    \par
    \caption{\textbf{Learning the diffusion-reaction system in porous media in Section~\ref{sec: PorousMedia}.} 
     (\textbf{A}) The training data includes a random $f_{\mathcal{T}}$ generated from GRF and the corresponding solution $C_{A,\mathcal{T}}$ and $C_{B,\mathcal{T}}$. 
     (\textbf{B}) The convergence of $L^2$ relative errors of FPI and cLOINN-random for various test cases of $C_A$ and $C_B$.
     (\textbf{C}) Prediction example of different approaches for a test case with $\sigma_{\text{test}} = 0.3$.}
    \label{fig:PorousSystem_res}
\end{figure}

Since there are two outputs for this case, we consider the local solution operator  (Fig.~\ref{fig:mesh}D)
\[\Tilde{\mathcal{G}}: 
\{C_A(x, t-\Delta t), C_A(x-\Delta x, t), C_A(x+\Delta x, t), C_B(x, t-\Delta t), C_B(x-\Delta x, t), C_B(x+\Delta x, t), f(x, t)\} \]
\[\mapsto (C_A(x, t), C_B(x, t)). \]
We show one example of training dataset $\mathcal{T}$ in Fig. \ref{fig:PorousSystem_res}A.

In this experiment, FPI works well, and the $L^2$ relative errors achieves less than 1\% when $\sigma_{\text{test}} = 0.10$. FPI performs better than cLOINN-random (Table~\ref{tab:L2error_more}). When $\sigma_{\text{test}} = 0.10, 0.30$, the accuracy of cLOINN-random is smaller than 10\%. It is shown that, in this example, FPI not only more accurate than cLOINN-random, but also converge faster (Fig.~\ref{fig:PorousSystem_res}B). We show examples of $C_A$ and $C_B$ prediction using FPI and cLOINN-random approaches in Fig.~\ref{fig:PorousSystem_res}C.


\section{Conclusions}
\label{sec:conclusion}

Learning solution operators of partial differential equations (PDEs) usually requires a large amount of training data. In this study, we propose, to the best of our knowledge, the first one-shot method to learn solution operators from only one PDE solution, which can potentially address the challenge of data acquisition in real applications. Considering small local domains instead of the entire computational domain of the PDE, we define and learn a local solution operator, which is then leveraged to predict a new solution via a fixed-point iteration (FPI) or local-solution-operator informed neural networks. Our method accommodates both unstructured data and grid data when training LOINN and cLOINN model.

The effectiveness of our method is demonstrated in various examples, including 1D/2D equations, linear/nonlinear equations, advection equation in which $f$ is an coefficient, diffusion-reaction system in porous media, and complex domain with a cutout. Among the proposed approaches, FPI and cLOINN-random are demonstrated to have comparable good performance across all test cases. The neural network-based approaches, LOINN and cLOINN, allow greater flexibility on complex domains and random point locations, with cLOINN generally outperforming LOINN in most situations and converges faster. Our experiments show that using randomly sampled point locations can improve the accuracy. Additionally, we show the effect of different local solution operators, and we find including more auxiliary points improves the performance. Our method has proven effective even on very coarse grids or a small number of point locations.

The limitation of our method lies in its diminished accuracy when predicting solutions divergent from the known solution, which is an inherent challenge arises from the constraints of utilizing minimal data. In the future, we will carry out further validation on PDEs with different boundary conditions or complex domains, and improve our approaches for better accuracy, faster convergence and better computational efficiency. Moreover, our method can potentially integrate with graph neural networks (GNN) \cite{sanchez2020learning, pfaff2020learning} to construct local graphs and train the local solution operator, and then combined it with FPI, LOINN or cLOINN for learning the solution operator.


\section*{Acknowledgments}

This work was supported by the U.S. Department of Energy [DE-SC0022953].

\appendix
\section{Hyperparameters}
\label{sec:hyperprarmeters}
The specific parameters for generating dataset for each example are listed in Table~\ref{tab:parameters}. The hyperparameters of the neural network of the local solution operators $\Tilde{\mathcal{G}}$ are listed in Table~\ref{tab:Hyperparameters}. The hyperparameters of neural networks in LOINN and cLOINN are listed in Table~\ref{tab:Hyper_loinn}.

\begin{table}[htbp]
    \footnotesize
    \caption{\textbf{Parameters of data generation for training $\Tilde{\mathcal{G}}$.} Parameters $l$ and $\sigma_{\text{train}}$ represent the length scale and amplitude of a Gaussian Random Field (GRF) from which the $f_{\text{random}}$ of the training data is sampled.}
    \label{tab:parameters}
    \centering
    \begin{tabular}{lccc}
    \toprule
    & $l$ & $\sigma_{\text{train}}$ & $\Delta x_1 = \Delta x_2$\\
    \midrule
    Section~\ref{sec: Poisson1D} 1D Poisson 
    & 0.01 & 0.5 & 0.01\\
    Section~\ref{sec: LinearDiffusion} 
    Linear diffusion & 0.01 & 0.1 & 0.01\\
    Section~\ref{sec: NonlinearDiffusion} 
    Nonlinear diffusion-reaction & 0.01 & 0.1 & 0.01, 0.02, 0.04, 0.05, 0.07\\
    Section~\ref{sec: Advection} Advection 
    & 0.01 & 0.5 & 0.01\\
    Section~\ref{sec: Poisson2D} 2D nonlinear Poisson 
    & 0.01 & 0.1 & 0.05\\
    Section~\ref{sec: Poisson2D_cutout} 2D nonlinear Poisson with a circle cutout 
    & 0.01 & 0.1 & 0.01 \\
    Section~\ref{sec: PorousMedia} Diffusion-reaction system in porous media 
    & 0.01 & 0.1 & 0.01\\
    \bottomrule
\end{tabular}
\end{table}

\begin{table}[htbp]
    \footnotesize
    \caption{\textbf{Hyperparameters of the neural network of the local solution operators $\Tilde{\mathcal{G}}$.} Adam and L-BFGS are used to train the nueral networks.}
    \label{tab:Hyperparameters}
    \centering
    \begin{tabular}{lccccc}
    \toprule
    &  Depth & Width & Adam iterations\\
    \midrule
    \multirow{1}{*}{Sec.~\ref{sec: Poisson1D} 1D Poisson}  
    & 2 & 64 & 200000 \\
    \multirow{1}{*}{Sec.~\ref{sec: LinearDiffusion} Linear diffusion}
    & 2 & 64 & 150000\\
    \multirow{1}{*}{Sec.~\ref{sec: NonlinearDiffusion} Nonlinear diffusion-reaction} 
    & 2 & 128 & 150000\\
    \multirow{1}{*}{Sec.~\ref{sec: Advection} Advection} 
    & 2 & 64 & 100000\\
    \multirow{1}{*}{Sec.~\ref{sec: Poisson2D} 2D nonlinear Poisson}  
    & 2 & 64 & 100000 \\
    \multirow{1}{*}{Sec.~\ref{sec: Poisson2D_cutout} 2D nonlinear Poisson with a circle cutout} 
    & 2 & 64 & 150000\\
    \multirow{1}{*}{Sec.~\ref{sec: PorousMedia} Diffusion-reaction system in porous media} 
    & 2 & 64 & 100000\\
    
    \bottomrule
\end{tabular}
\end{table}

\begin{table}[htbp]
    \footnotesize
    \caption{\textbf{Hyperparameters of the neural networks in LOINN and cLOINN.} Adam is used to train the neural networks.}
    \label{tab:Hyper_loinn}
    \centering
    \begin{tabular}{lccccc}
    \toprule
    &  Depth & Width & Iterations\\
    \midrule
    \multirow{1}{*}{Sec.~\ref{sec: Poisson1D} 1D Poisson}  
    & 3 & 128 & 20000 \\
    \multirow{1}{*}{Sec.~\ref{sec: LinearDiffusion} Linear diffusion}
    & 3 & 128 & 100000\\
    \multirow{1}{*}{Sec.~\ref{sec: NonlinearDiffusion} Nonlinear diffusion-reaction} 
    & 3 & 64 & 100000\\
    \multirow{1}{*}{Sec.~\ref{sec: Advection} Advection} 
    & 3 & 64 & 100000\\
    \multirow{1}{*}{Sec.~\ref{sec: Poisson2D} 2D nonlinear Poisson}  
    & 2 & 16 & 20000\\
    \multirow{1}{*}{Sec.~\ref{sec: Poisson2D_cutout} 2D nonlinear Poisson with a circle cutout} 
    & 3 & 128 & 150000\\
    \multirow{1}{*}{Sec.~\ref{sec: PorousMedia} Diffusion-reaction system in porous media} 
    & 3 & 32 & 30000\\
    
    \bottomrule
\end{tabular}
\end{table}

\section{Reasoning of using neural networks for linear and nonlinear PDEs}
\label{sec:nn_reason}

The foundation of our method is that the same PDE is satisfied in an arbitrary-shaped small domain $\tilde{\Omega}$ inside the entire domain $\Omega$. Based on this fact, we define and train a local solution operator $\tilde{\mathcal{G}}$ at $\tilde{\Omega}$, which captures the local relationship dictated by the PDE. In this section, we demonstrate the reason of using neural network as a local solution operator.

For a PDE we aim to solve, there are three cases. 
\begin{enumerate}
    \item We do not know if the PDE is linear or nonlinear. In this case, we use neural networks for $\tilde{\mathcal{G}}$ as a universal approximator.
    \item If we know that the PDE is nonlinear, we use the neural network $\tilde{\mathcal{G}}$ to capture the nonlinear relationship of nodes in the local domain.
    \item If we know that the PDE is linear, we could use neural networks or simply the linear regression. However, we will show that neural networks can outperform linear regression models in capturing these relationships in the following. 
\end{enumerate} 

\subsection{1D Poisson equation}

We use the 1D Poisson equation in Section~\ref{sec: Poisson1D} to demonstrate that neural networks outperform linear regression models even for linear PDEs.

In our method, we choose the simplest local solution operator $\tilde{\mathcal{G}}$ using 3 nodes with $\Delta x = 0.01$ in Fig.~\ref{fig:mesh}B. We construct the training dataset for $\tilde{\mathcal{G}}$ by generating the input-output pairs 
$$\Bigl\{\underbrace{\Bigl[u(x - \Delta x), u(x + \Delta x),f(x)\Bigl]}_{\text{Input}}, \underbrace{u(x)}_{\text{Output}}\Bigl\}_{x}.$$  We explore two choices for learning the local solution operator $\tilde{\mathcal{G}}$: a linear regression model and a neural network.

We start by using a linear regression (LR) model to predict the solution:
\[ 
\hat{u}(x) = \beta_1  u(x - \Delta x) + \beta_2  u(x + \Delta x) + \beta_3 f(x) + \beta_0, 
\]
where \(\beta_1\), \(\beta_2\), and \(\beta_3\) are the coefficients learned through the training process to best fit the data. As the inclusion of the bias term does not improve performance, we trained the LR model without \(\beta_0\). The resulting coefficients \(\beta_1\), \(\beta_2\), and \(\beta_3\) are 0.5003, 0.4998, and -0.0047, respectively, so the linear relationship is 
\[\hat{u}(x) = 0.5003 u(x - \Delta x) + 0.4998 u(x + \Delta x) - 0.0047 f(x) \]
\[\approx \frac{1}{2} \big(u(x - \Delta x) + u(x + \Delta x) - 100(\Delta x)^2f(x)\big). \]
The last equation is obtained by applying the central difference formula for the second-order derivative to the PDE. We observe that there is only a slight difference between the coefficients from LR model and finite difference.

\begin{figure}[htbp]
    \centering
    \includegraphics[width=\textwidth]{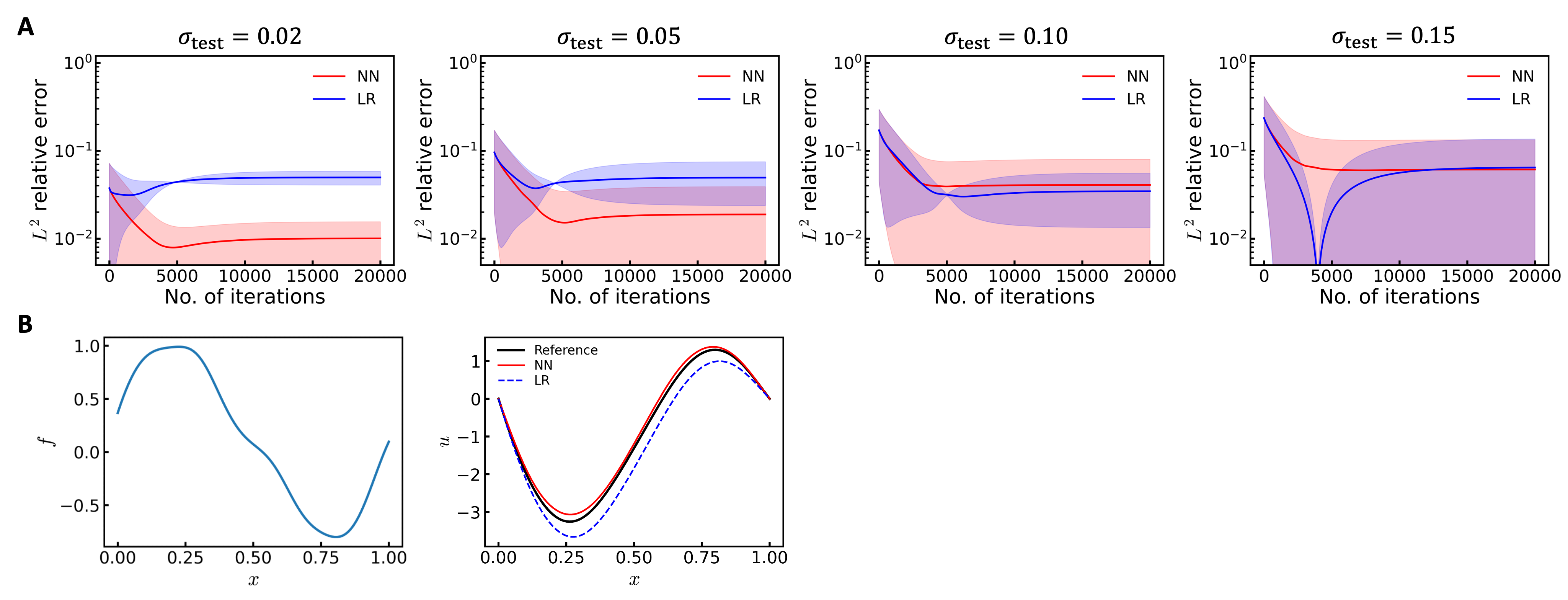}
    \par
    \caption{\textbf{Learning the 1D Poisson equation in Section~\ref{sec: Poisson1D}.}
    (\textbf{A}) The comparison of the convergence of $L^2$ relative errors between LR model and NN model as local solution operators for various test cases using FPI approach.
     (\textbf{B}) One example of the new $f$ and the predictions of FPI approach using LR model and NN model as local solution operators for the test case with $\sigma_{\text{test}} = 0.15$.}
    \label{fig:Poisson_analysis}
\end{figure}

We also consider the neural network (NN) as a local solution operator shown in Section~\ref{sec: Poisson1D}. We test both LR model and NN model as local solution operators for various cases with different $\sigma_{\text{test}}$ utilizing FPI approach (Fig.~\ref{fig:Poisson_analysis}A). The neural network $\tilde{\mathcal{G}}$ achieves better accuracy for most cases, and we observe that it has a larger ratio of success trials than LR model. For a new $f$, we show an example of the predictions (Fig.~\ref{fig:Poisson_analysis}B). For the NN, it learns this relationship with better accuracy than LR model.  

From the discussion above, we show that the ``locality" can be captured by neural networks in a linear system more effectively than a linear regression model. Note that generally we do not know the exact form of the equation, and neural network is a preferred choice of the local solution operator $\tilde{\mathcal{G}}$ in our method. 

\bibliographystyle{unsrt}
\bibliography{main}

\end{document}